\pdfoutput=1 % for arxiv submission
\documentclass[journal]{IEEEtran}
\usepackage[noadjust]{cite}
\ifCLASSINFOpdf
  \usepackage[pdftex]{graphicx}
\else
  \usepackage[dvips]{graphicx}
\fi
\usepackage{amsmath}
\usepackage{array}
\ifCLASSOPTIONcompsoc
  \usepackage[caption=false,font=normalsize,labelfont=sf,textfont=sf]{subfig}
\else
  \usepackage[caption=false,font=footnotesize]{subfig}
\fi
\usepackage{url}
\usepackage{threeparttable}

\begin{document}
%
% paper title
% Titles are generally capitalized except for words such as a, an, and, as,
% at, but, by, for, in, nor, of, on, or, the, to and up, which are usually
% not capitalized unless they are the first or last word of the title.
% Linebreaks \\ can be used within to get better formatting as desired.
% Do not put math or special symbols in the title.
\title{Accelerator-Aware Pruning for Convolutional Neural Networks}
%
%
% author names and IEEE memberships
% note positions of commas and nonbreaking spaces ( ~ ) LaTeX will not break
% a structure at a ~ so this keeps an author's name from being broken across
% two lines.
% use \thanks{} to gain access to the first footnote area
% a separate \thanks must be used for each paragraph as LaTeX2e's \thanks
% was not built to handle multiple paragraphs
%

\author{Hyeong-Ju~Kang,~\IEEEmembership{Member,~IEEE}%
%\author{Hyeong-Ju~Kang,~\IEEEmembership{Member,~IEEE,}
%        and~Byung-Do~Yang%
\thanks{Manuscript received XXXXX, XX, XXXX; revised XXXXX XX, XXXX.
This work was supported by Basic Science Research Program
through the National Research Foundation of Korea (NRF)
funded by the Ministry of Education(2015R1D1A1A01058768).
This work was also supported by IDEC (EDA Tool).}%
\thanks{H.-J. Kang is with the School of Computer Science and Engineering,
	Korea University of Technology and Education, Cheonan, Chungnam,
	31253 Republic of Korea e-mail: hjkang@koreatech.ac.kr.}%
%\thanks{B-D. Yang is with the School of Electronics Engineering,
%	Chungbuk National University, Cheongju, Chungbuk,
%	28644 Republic of Korea e-mail: bdyang@cbnu.ac.kr.}%
}

% note the % following the last \IEEEmembership and also \thanks - 
% these prevent an unwanted space from occurring between the last author name
% and the end of the author line. i.e., if you had this:
% 
% \author{....lastname \thanks{...} \thanks{...} }
%                     ^------------^------------^----Do not want these spaces!
%
% a space would be appended to the last name and could cause every name on that
% line to be shifted left slightly. This is one of those "LaTeX things". For
% instance, "\textbf{A} \textbf{B}" will typeset as "A B" not "AB". To get
% "AB" then you have to do: "\textbf{A}\textbf{B}"
% \thanks is no different in this regard, so shield the last } of each \thanks
% that ends a line with a % and do not let a space in before the next \thanks.
% Spaces after \IEEEmembership other than the last one are OK (and needed) as
% you are supposed to have spaces between the names. For what it is worth,
% this is a minor point as most people would not even notice if the said evil
% space somehow managed to creep in.

% The paper headers
\markboth{}%
{Shell \MakeLowercase{\textit{et al.}}: Bare Demo of IEEEtran.cls for IEEE Journals}
% The only time the second header will appear is for the odd numbered pages
% after the title page when using the twoside option.
% 
% *** Note that you probably will NOT want to include the author's ***
% *** name in the headers of peer review papers.                   ***
% You can use \ifCLASSOPTIONpeerreview for conditional compilation here if
% you desire.

% If you want to put a publisher's ID mark on the page you can do it like
% this:
%\IEEEpubid{0000--0000/00\$00.00~\copyright~2015 IEEE}
% Remember, if you use this you must call \IEEEpubidadjcol in the second
% column for its text to clear the IEEEpubid mark.

% use for special paper notices
%\IEEEspecialpapernotice{(Invited Paper)}

% make the title area
\maketitle

% As a general rule, do not put math, special symbols or citations
% in the abstract or keywords.
\begin{abstract}
Convolutional neural networks have shown tremendous performance capabilities
	in computer vision tasks,
	but their excessive amounts of weight storage and arithmetic operations
		prevent them from being adopted in embedded environments.
One of the solutions involves pruning, where certain unimportant weights
	are forced to have a value of zero.
Many pruning schemes have been proposed,
	but these have mainly focused on the number of pruned weights,
	scarcely considering ASIC or FPGA accelerator
		architectures.
When a pruned network is run on an accelerator,
	the lack of the architecture consideration causes
		some inefficiency problems,
	including internal buffer misalignments and load imbalances.
This paper proposes a new pruning scheme
	that reflects accelerator architectures.
In the proposed scheme, pruning is performed
	so that the same number of weights remain for each weight group 
		corresponding to activations fetched simultaneously.
In this way, the pruning scheme resolves the inefficiency problems,
	doubling the accelerator performance.
Even with this constraint,
	the proposed pruning scheme reached a pruning ratio similar
		to that of previous unconstrained pruning schemes,
		not only on AlexNet and VGG16 but also
	on state-of-the-art very deep networks such as ResNet.
Furthermore, the proposed scheme demonstrated a comparable pruning ratio
	on compact networks such as MobileNet and on slimmed networks
		that were already pruned in a channel-wise manner.
In addition to improving the efficiency of previous sparse accelerators,
	it will be also shown that the proposed pruning scheme can be used to reduce
		the logic complexity of sparse accelerators.
The pruned models are publicly available at
	\url{https://github.com/HyeongjuKang/accelerator-aware-pruning}
\end{abstract}

% Note that keywords are not normally used for peerreview papers.

% For peer review papers, you can put extra information on the cover
% page as needed:
% \ifCLASSOPTIONpeerreview
% \begin{center} \bfseries EDICS Category: 3-BBND \end{center}
% \fi
%
% For peerreview papers, this IEEEtran command inserts a page break and
% creates the second title. It will be ignored for other modes.
\IEEEpeerreviewmaketitle

\newenvironment{processtable}[3]{
	\caption{#1}
	#2		% label
	\centering
	#3
}{}
\newcommand{\reffig}[1]{Fig.~\ref{#1}}
\newcommand{\reftab}[1]{Table~\ref{#1}}
\newcommand{\refeqn}[1]{(\ref{#1})}

\def\form{2}
\section{Introduction}
\IEEEPARstart{C}{onvolutional} neural networks (CNNs) are attracting
	interest in the fields of image recognition
		\cite{krizhevsky12, simonyan15, szegedy15, he16, iandola16},
	object detection
		\cite{sermanet14, girshick14, girshick15, ren15_2, mclaughlin17},
	and image segmentation \cite{long15}.
Although they provide great performance for computer vision tasks,
	there are certain obstacles that must be overcome before they can be adopted 
		in embedded environments.
A CNN usually requires excessive amounts of weight storage
	and arithmetic operations.
For fast processing and low power consumption under these requirements,
	ASIC or FPGA accelerators have been proposed
	\cite{chen14, albericio16, zhang15, song16, qiu16, motamedi16, jo18},
but the amounts of weights and operations remain a major concern.

The weight amounts can be reduced by network pruning
	\cite{han15, han16, yang17
	, wen16, li17, he17, yu17_2, molchanov17
	, yu17, lebedev16, anwar17, mao17, kadetotad16, park17, boo17},
%figurnov16: perforated CNN
	where some unimportant weights are forced to have a value of zero.
Multiplication with zero is meaningless,
	implying that reductions in operation amounts can be expected as well.
Various pruning schemes have been proposed,
	some of which prune the weights without constraints \cite{han15, han16}
	while others prune the weights considering the neural network structures.
It is known that relatively more weights can be pruned 
	in fully-connected layers than in convolutional layers.
However, pruning convolutional layers can reduce
	more energy and realize higher throughput \cite{yang17}.
Given that the operational structure of convolutional layers
	is more complex than that of fully-connected layers,
	convolutional layers have a greater variety of pruning schemes 
	\cite{wen16, li17, he17, yu17_2, molchanov17
		, lebedev16, yu17, anwar17, kadetotad16, mao17, park17}.
%figurnov16: perforated CNN

Pruned networks can be executed on certain ASIC or FPGA accelerators
	that can exploit the weight sparsity.
The EIE proposed in \cite{han16_2} used the sparsity of weights 
	generated by the pruning algorithm of \cite{han16},
	but the architecture only processes fully-connected layers.
The EIE architecture was modified for a long short-term memory (LSTM) network
	to become the ESE architecture \cite{han17}.
Cambricon-X is an architecture that can exploit weight sparsity
	in convolutional layers \cite{zhang16_2}.
SCNN proposed in \cite{parashar17} exploits both weight sparsity
	and activation sparsity for convolutional layers.
In ZENA proposed in \cite{kim17}, good operating efficiency was reached
	by skipping zero weights and activations.

Despite the various pruning schemes and accelerators,
	it remains challenging to utilize the performance
	of ASIC or FPGA accelerators efficiently with pruned networks.
The aforementioned pruning schemes were developed 
	without considering the concept of accelerator acceptability.
They mainly focused on the amount of weight that can be pruned away.
Pruning with no constraints creates irregular patterns
	in the remaining non-zero weights.
These irregular patterns cause some degree of inefficiency
	when the pruned networks are performed on accelerators.
The misalignment of the fetched activations and weights
	requires padding-zero insertions.
The processing elements (PEs) process different numbers of weights,
	meaning that some PEs must wait for other PEs to complete,
		known as the load-imbalance problem.
To alleviate these problems, some accelerators use complex structures.
For example, Cambricon-X uses very wide (256$\times$16-bits width)
	memory and very wide (256-to-1) multiplexers (MUXs).

This paper proposes a new approach,
	a pruning algorithm which considers the accelerator architecture.
There can be various architecture consideration points, and
	this paper will focus on the activation groups fetched and processed
			simultaneously in a PE
		and the number of remaining weights for each group.
These points are related to certain critical parameters in accelerator designs,
	including the number of multipliers
		and the width of the internal weight buffer.
In the proposed algorithm, pruning is applied to weight groups, 
	each of which corresponds to an activation group fetched together, and
	after pruning, a fixed number of weights remain in a group.
Because pruning is applied to a weight group aligned
	with the operating boundary of a PE,
	the scheme can resolve the misalignment problem,
		removing waste of the internal buffer or multipliers.
The pruning group can be adjusted
	to reduce the complexity of the data selection and indexing logic.
Furthermore, since the remaining non-zero weights are distributed evenly,
	the load-imbalance problem can be solved naturally.
	
This paper is organized as follows.
Section II explains CNNs, the previous pruning schemes,
	and the accelerator architectures.
The accelerator-aware pruning scheme is proposed in Section III,
	and the experimental results are shown in Section IV.
Section V makes the concluding remarks.

\section{CNN and Pruning}

\subsection{CNN}
A CNN usually consists of many convolutional layers
	and a few fully-connected layers.
Between the layers, there are activation layers
		such as rectified linear units (ReLUs),
	pooling layers, and batch-normalization layers.
It is known that the convolutional layers account for
	more than 90\% of the arithmetic operations.
Because the fully-connected layers require much more weight storage,
	recent CNNs have tended to have
		only one fully-connected layer \cite{szegedy15, he16}
		or none \cite{iandola16}.

\begin{figure}[!t]
	\centering
	\includegraphics[scale=0.5]{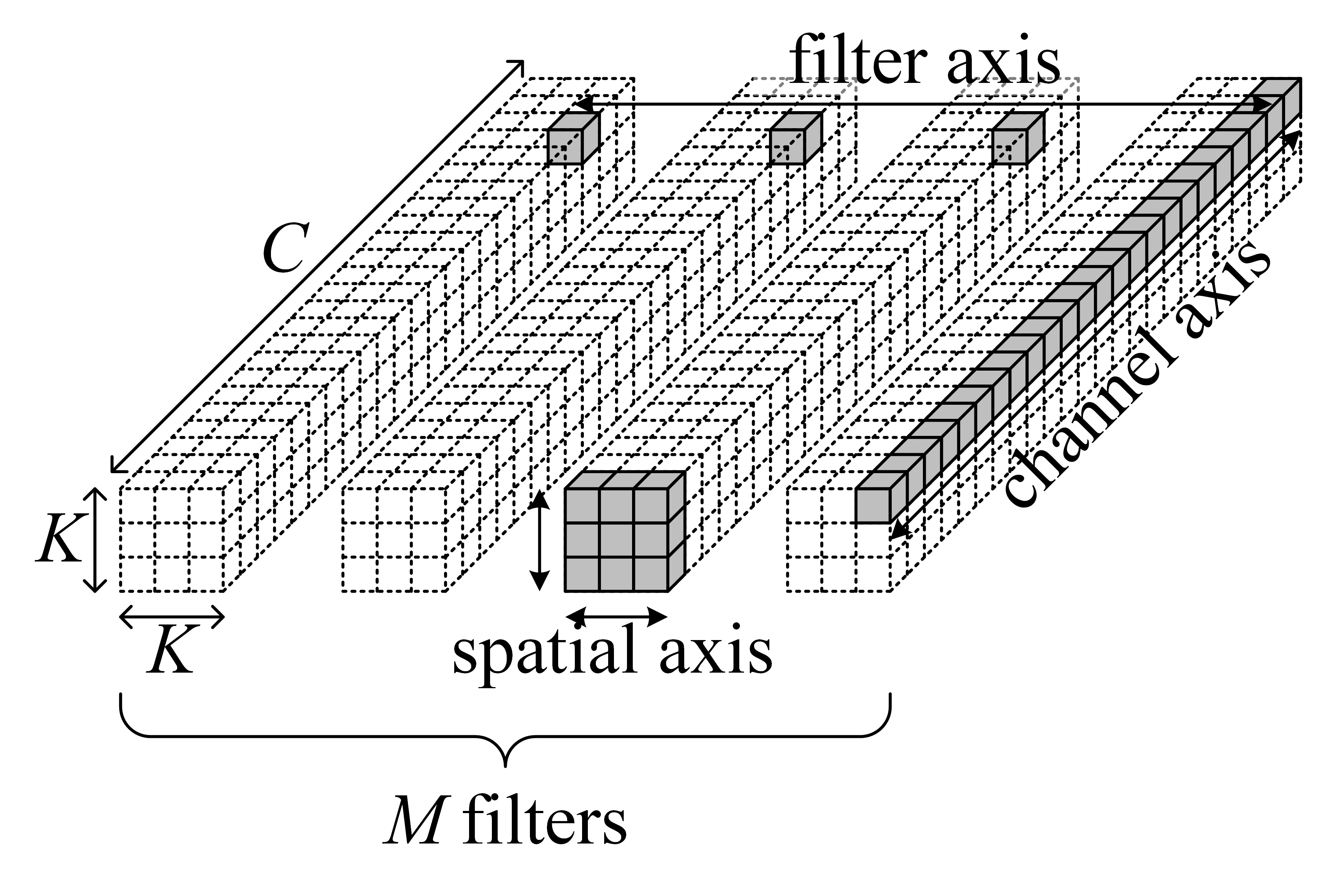}
	\caption{Weight structure of a convolutional layer.}
	\label{fig:axis}%
\end{figure}

The operation in a convolutional layer is as follows:
\begin{align}
fo(m,y,x) = \sum^{C-1}_{c=0} \sum^{K-1}_{i=0} \sum^{K-1}_{j=0}
				weight(m,c,i,j) \times\\
			fi(c,S \times y + i, S \times x + j) + bias(m),	\nonumber
\label{eqn:conv}
\end{align}
where $fo(m,y,x)$ is the activation of row $y$ and column $x$
		in output feature map $m$
	, and $fi(c,h,w)$ is the activation of row $h$ and column $w$
		in input feature map $c$.
In the equation, $C$ is the number of the input feature maps or channels,
	$K$ is the spatial size of the kernel, and $S$ is the stride size.
The $weight(m,c,i,j)$ variables are the convolution weights,
	and the $bias(m)$ variables are the bias terms.
\reffig{fig:axis} shows the structure of the weights,
	consisting of $M$ filters with a spatial size of $K{\times}K$ 
	and a depth of $C$.

For ease of discussion, some axes are defined.
The channel axis of the input activations denotes the $fi(c,h,w)$ variables
	with identical values of $h$ and $w$.
Similarly, the channel axis of the weights is defined
	by the $weight(m,c,i,j)$ entries with the same $m$, $i$, and $j$ values.
The spatial axis represents $fi(c,h,w)$ variables with the same $c$ values
	or $weight(m,c,i,j)$ variables with identical values of $m$ and $c$ as well.
The filter axis denotes $weight(m,c,i,j)$ variables
	with the same  $c$, $i$, and $j$ values.
\reffig{fig:axis} shows the weights along each axis.

\subsection{Pruning}

\begin{figure*}[!t]
	\centering
	\subfloat[]{
		\includegraphics[scale=0.7]{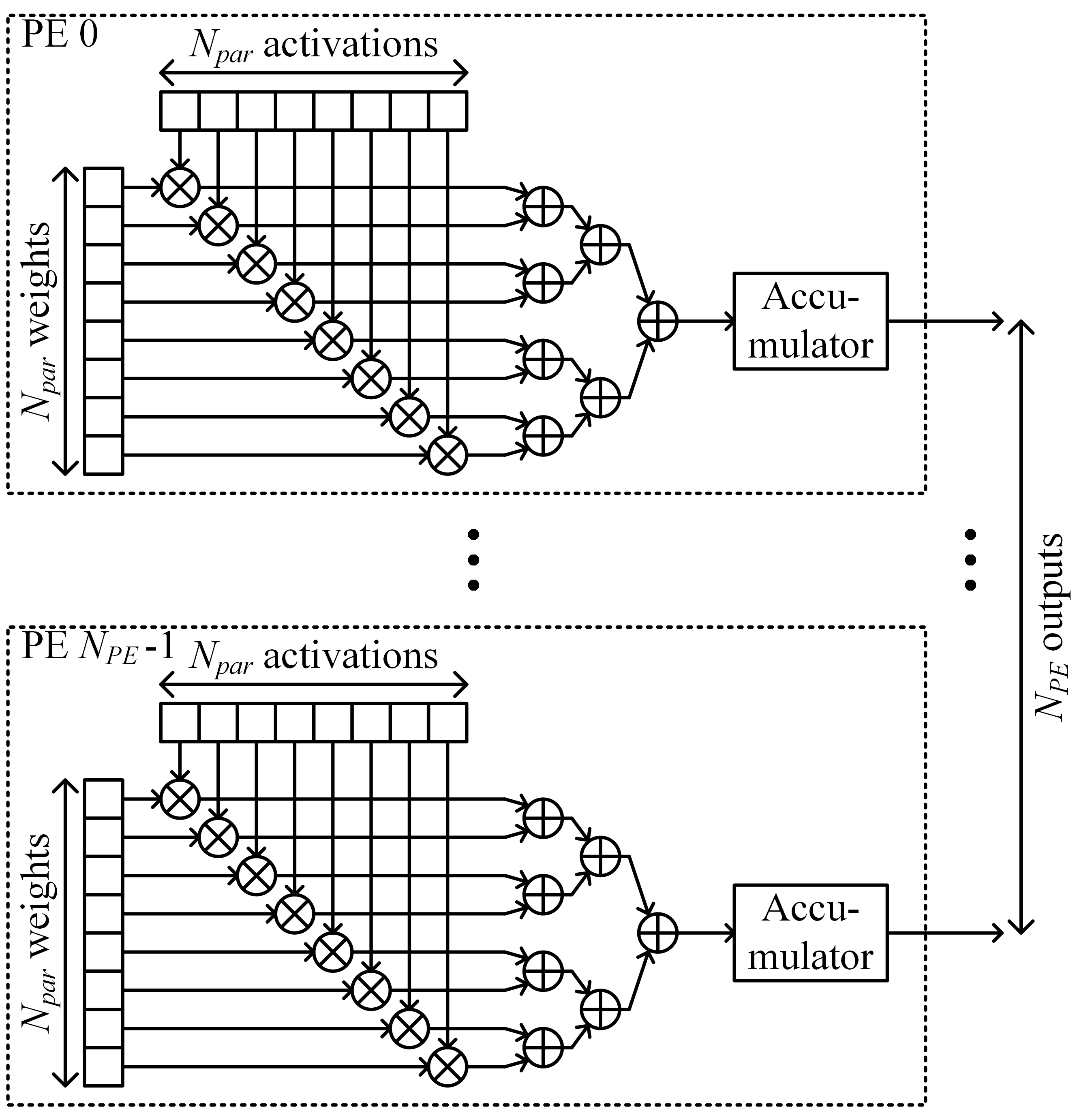}%
		\label{fig:pe_mwma}%
	}~~~%
	\hfil
	\subfloat[]{
		\includegraphics[scale=0.7]{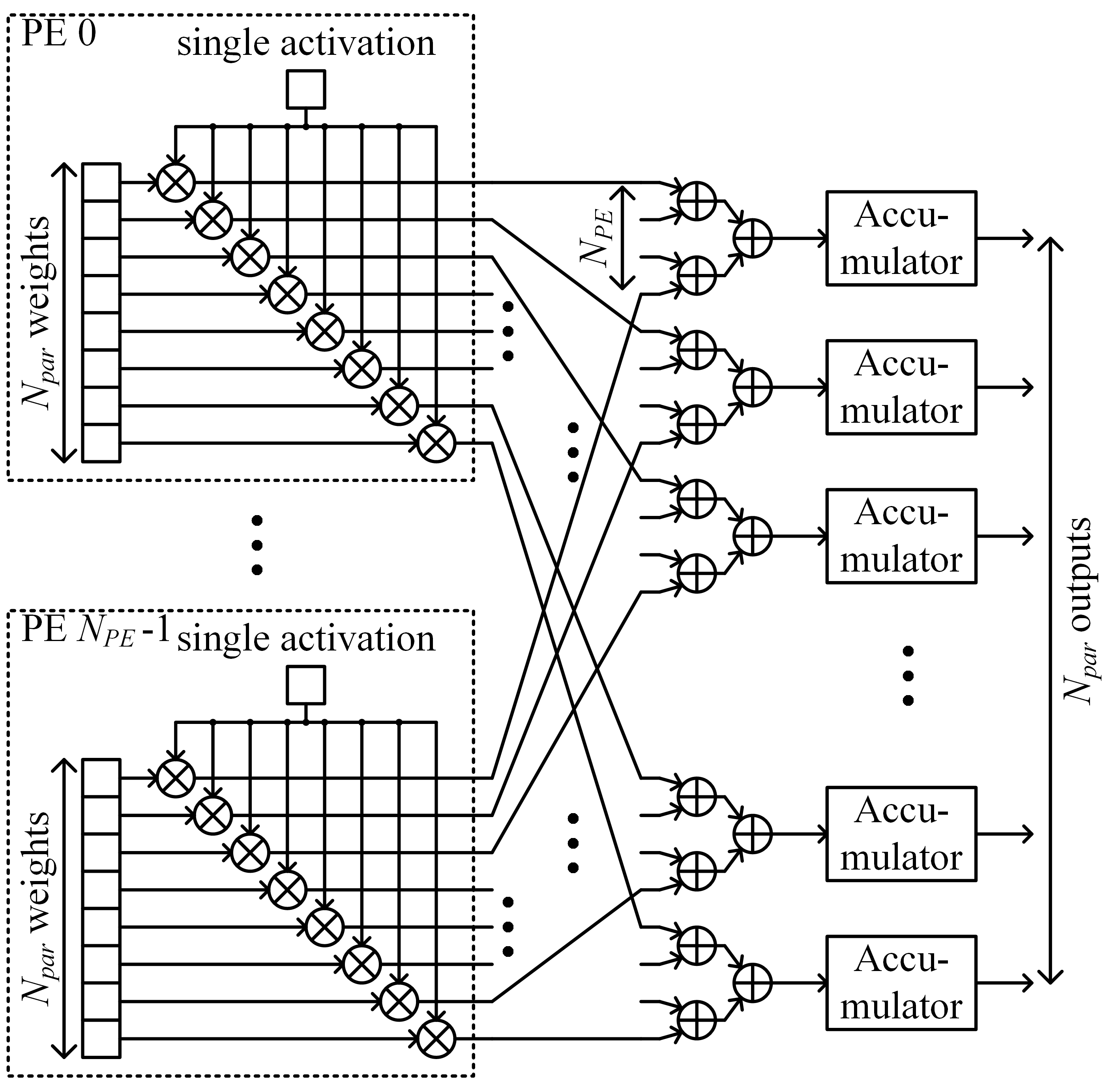}%
		\label{fig:pe_mwsa}%
	}%
	\caption{Typical PE structures: (a) MWMA and (b) MWSA.
	}
	\label{fig:pe}
\end{figure*}

\begin{table}[!t]
	\processtable{Pruning ratio of the previous pruning scheme in \cite{han15}}
	{\label{tab:prev1}}
	{
\begin{tabular}{ l | c c c c }
	\hline
	 & Conv1 & Conv & FC1\&FC2 & FC3 \\
	\hline
	\hline
	AlexNet & 16\% & 62--65\% & 91\% & 75\% \\
	VGG16   & 42\% & 47--78\% & 96\% & 77\% \\
	\hline
\end{tabular}

	} {}
\end{table}

The pruning of neural networks removes some of the unimportant weights or nodes
	to reduce the amount of storage and the number of operations.
The early works were presented in \cite{lecun90, hassibi93_2, hassibi93},
	and pruning in CNNs was proposed in \cite{han15},
	whose pruning results are summarized in \reftab{tab:prev1}.
The table shows the ratio of the pruned weights
	at the first convolutional layer, the other convolutional layers,
	the first and second fully-connected layers, and the last fully-connected
	layer of AlexNet and VGG16.
This work was expanded with quantization and Huffman coding in \cite{han16}
	and with an accelerator architecture for pruned fully-connected layers 
		in \cite{han16_2}.
The pruning scheme can prune many weights but shows irregularities
	in the pruned pattern.
Moreover, the corresponding accelerator architecture can only deal with
		fully-connected layers.
The energy-aware pruning scheme proposed in \cite{yang17} 
	focuses on convolutional layers because convolutional layers consume
		more energy.
However, this work did not consider the regularity of the pruning pattern,
	either.

\begin{table}[!t]
	\processtable{Previous structured pruning schemes}
	{\label{tab:prev_prune}}
	{

\begin{tabular}{ l | l | l}
	\hline
	Granularity & Pruned weights & References\\
	\hline
	\hline
	channel-wise or & $weight(:,n,:,:)$ or & \cite{wen16, yu17, li17,
									he17, yu17_2, molchanov17}\\
	filter-wise & $weight(m,:,:,:)$ & \\
	\hline
	shape-wise or GPU-aware & $weight(:,n,i,j)$ & \cite{lebedev16,anwar17} \\
	\hline
				  	& $weight(m,n,:,:)$ & \\
	2D/1D granularity & $weight(m,n,i,:)$ & \cite{mao17}\\
					& $weight(m,n,:,j)$ &  \\
	\hline
	SIMD aware & & \cite{yu17_2} \\
	\hline
\end{tabular}

	} {}
\end{table}

The regularity of the pruning was considered in
	\cite{wen16, yu17, li17, he17, yu17_2, molchanov17, lebedev16, anwar17
		, mao17, kadetotad16}.
The pruning schemes in these studies can be categorized as
	channel-wise, filter-wise, and shape-wise pruning,
	as shown in \reftab{tab:prev_prune}.
In channel-wise pruning, for example, 
	all of the weights in a channel are pruned or not together.
%In \cite{mao17}, various level of granularity.
%	0-D: fine-grained.
%	1-D: Sub-kernel vector in a 2-D kernel.
%	2-D: a 2-D kernel.
%	3-D: filter-wise pruning.
These pruning schemes are referred to as structured pruning schemes,
	whereas pruning schemes with no constraint,
			as presented in \cite{han15, han16},
		are termed unstructured pruning.

%Some pruning schemes target a General-Purpose Graphic Processing Unit (GPGPU).
%In a GPGPU implementation, the convolution is usually transformed
%	into a matrix multiplication.
%By removing columns in the transformed weight matrix,
%	the amount of operations can be reduced \cite{lebedev16}.
%A column in the weight matrix corresponds to the weights, $weight(:,n,i,j)$'s,
%	so this scheme can be thought of as a shape-wise pruning.
%A strided version was proposed in \cite{anwar17}.

\subsection{Neural Network Accelerators}
Some ASIC or FPGA accelerator architectures have been
	proposed for unpruned, dense CNNs,
		which will be called dense architectures.
An accelerator usually consists of several processing elements (PEs),
	each of which has a single multiplier or many multipliers.
In a PE with many multipliers,
	the multipliers may multiply
	multiple weights and multiple activations (MWMA),
	multiple weights and a single activation (MWSA),
	or a single weight and multiple activations (SWMA).
The MWMA and MWSA structures are shown in \reffig{fig:pe},
	where $N_{PE}$ PEs operate in parallel.
With the same naming convention, a PE with a single multiplier
	can be called a single weight and single activation (SWSA) structure.

If a PE has multiple multipliers,
	it is important to fetch the operands of the multipliers simultaneously.
In this paper, a fetching group is defined as activations or weights
	that are fetched and processed simultaneously in a PE.
If the size of a fetching group is $N_{par}$
	and the number of multipliers in a PE is $N_{mul}$,
	$N_{par}$ is usually equal to $N_{mul}$.
The weights and activations are usually fetched from the internal buffers,
	but the buffers are not drawn in the figures.

%\begin{figure}[!t]
%	\centering
%	\includegraphics[scale=0.45]{figures/prev2.png}
%	\caption{Process of a channel-axis-parallel CNN accelerator.}
%	\label{fig:prev2}%
%\end{figure}

The PE structures can be further categorized by the axis
	followed by the weight- and activation-fetching groups.
For example, DianNao\cite{chen14} adopts the MWMA structure,
	where some of the input activations along the channel axis 
		are fetched and multiplied
			with the corresponding kernel weights.
The multiplication results are summed and accumulated 
	to be an output activation.
%A process example of such an architecture is shown in \reffig{fig:prev2}.
The MWSA structure is adopted in Cnvlutin \cite{albericio16}
	to exploit the sparsity of the input activations.
Multiple weights are fetched along the filter axis.
%The Cnvlutin architecture skips the MAC operations
%	of zero valued input activations.
The outputs of the multiplications in PEs are gathered
	into $N_{par}$ $N_{PE}$-input adder trees.
The architectures proposed in \cite{qiu16, motamedi16}
	consist of MWMA-structured PEs
	and fetch the activations and weights along the spatial axis.

\subsection{Sparse Accelerator Architectures}

\begin{figure}[!t]
	\centering
	\includegraphics[scale=0.75]{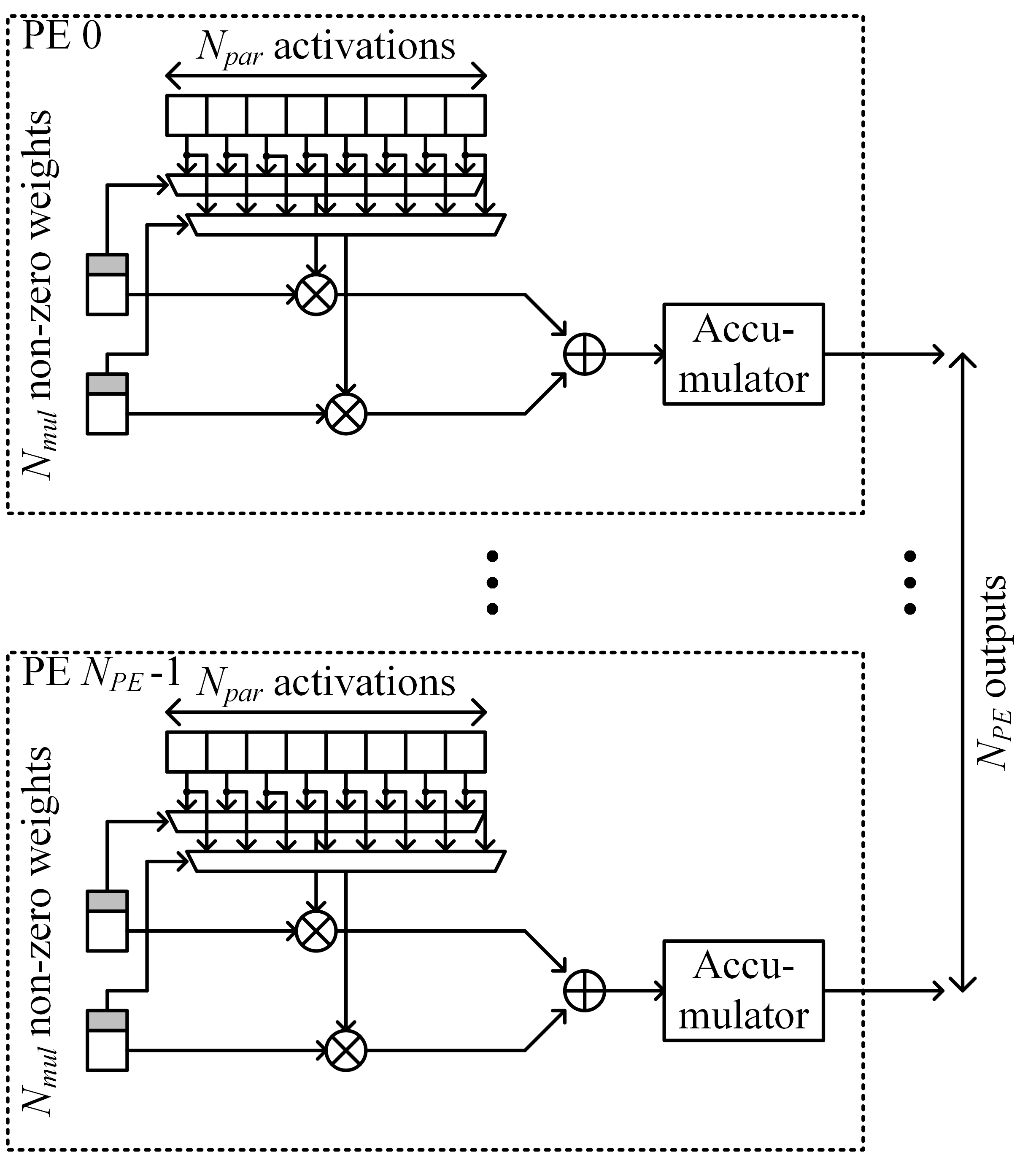}%
	\caption{Sparse MWMA PE structure.}
	\label{fig:pe_mwma_sparse}%
\end{figure}

\begin{figure}[!t]
	\centering
	\includegraphics[scale=0.45]{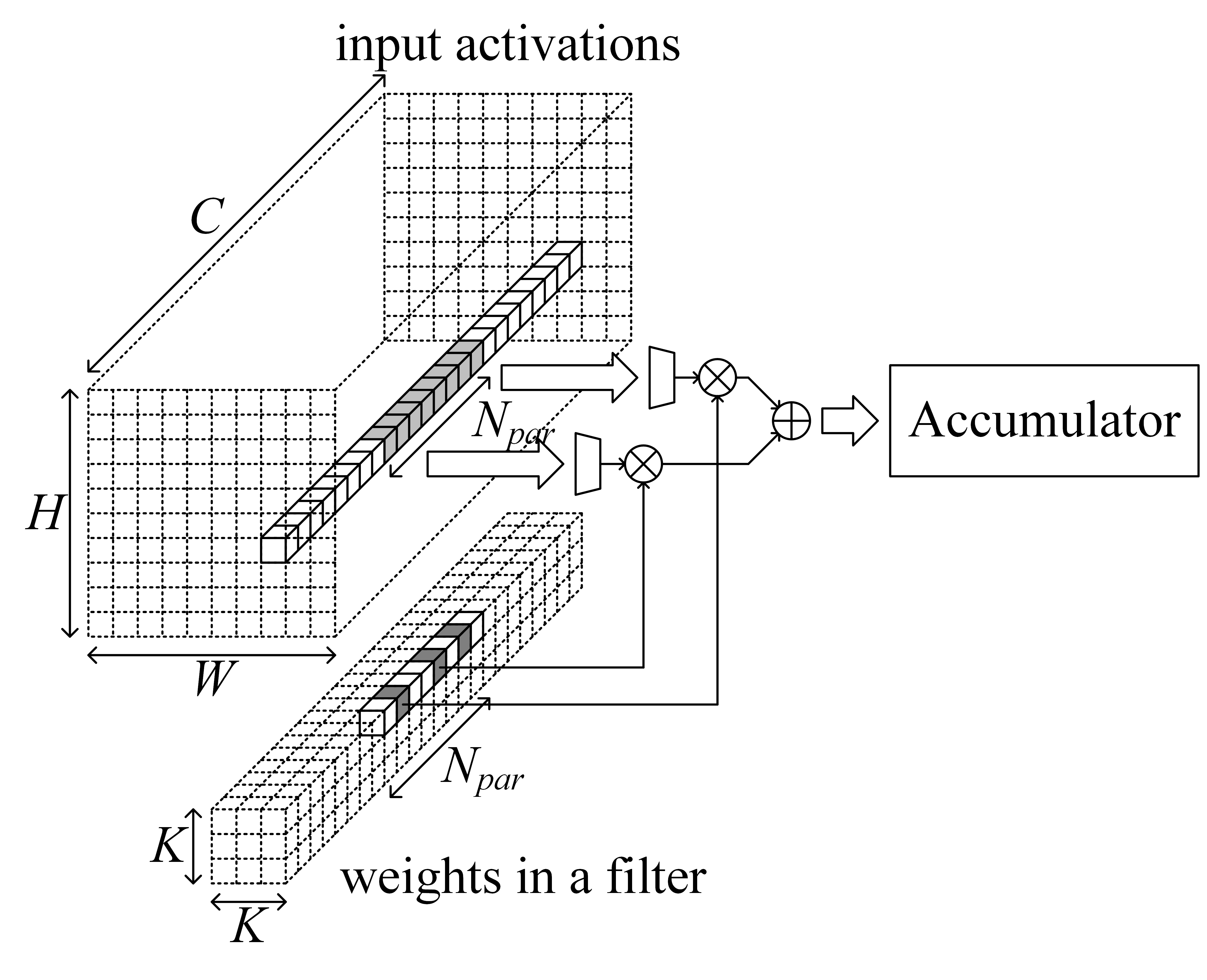}
	\caption{Process of a sparse channel-axis-parallel CNN accelerator.}
	\label{fig:sparse_ex}%
\end{figure}

The architectures in \reffig{fig:pe} can be modified
	to exploit the weight sparsity,
		which will be called sparse architectures.
In such architectures, only non-zero weights are stored in the weight memories
	to reduce weight storage use.
An example of the sparse MWMA structure
		is shown in \reffig{fig:pe_mwma_sparse},
	where $N_{par}$ activations are fetched simultaneously.
$N_{par}$ is usually larger than the number of multipliers, $N_{mul}$,
	as some weights are zero
	and multiplications with zero weights are meaningless.
Multipliers receive non-zero weights and their corresponding activations.
Selecting the $N_{mul}$ activations from the $N_{par}$ fetched activations
	requires $N_{mul}{\cdot}b$ $N_{par}$-to-1 MUXs,
		where $b$ is the bit width of the activations.
To select a proper activation, a non-zero weight is stored with an index,
	denoted by the grey rectangle in \reffig{fig:pe_mwma_sparse}.
A process example of the sparse MWMA PE structure is illustrated
	in \reffig{fig:sparse_ex},
	where non-zero weights are shown in grey.
In the figure, the upper MUX selects the second activation from the front,
	and the lower MUX selects the fifth activation.

As an example of such sparse architectures,
	Cambricon-X consists of
		the sparse MWMA structure PEs with $N_{par}$=256 and $N_{mul}$=16.
In this architecture, 256 activations are fetched simultaneously,
	from which 16 activations are selected.
For the selection, there are 16${\cdot}b$ 256-to-1 MUXs for each PE,
	and the MUXs are gathered into the indexing module (IM).

For ease of discussion,
	the weight-fetching group in a sparse architecture
	is defined to include all of the zero and
		non-zero weights corresponding to an activation-fetching group.
%If the number of non-zero weights in a weight fetching group
%	is larger than $N_{mul}$,
%	the weights are fetched and processed through several cycles
%		with the same activation group.
If the number of non-zero weights in a weight-fetching group
	is $N_{non-zero}$,
	it takes $\lceil N_{non-zero}/N_{mul} \rceil$ cycles
		for a PE to process the activation- and weight-fetching group.
In \reffig{fig:sparse_ex}, one more cycle is required
	to process the third non-zero weight.

Other PE structures have been used in sparse accelerators as well.
EIE \cite{han16_2}, ESE \cite{han17}, and ZENA \cite{kim17} 
	adopt SWSA structured PEs
		to easily skip the pruned weights in the irregular pattern.
% for sparse fully-connected layers.
%To exploit the irregular pattern of sparse weights,
%	they exploit SWSA structured PEs.
%ZENA \cite{kim17} also adopts SWSA structured PEs
%	and skips both of zero-valued activations and weights.
SCNN \cite{parashar17} has multiple multipliers in a PE
	but exploits a special structure of Cartesian Products.
In this structure, all of the fetched non-zero activations are multiplied
	with all of the fetched non-zero weights.
The multiplication results are delivered to proper output accumulators.

\subsection{Previous Pruning Scheme and Accelerator Architecture}
\label{subsec:prev}
Previous unstructured pruning schemes could reach high pruning ratios,
	but they rarely resulted in pruned networks that
		fit sparse accelerator architectures well.
The main reason for this is the non-uniform distribution
		of the non-zero weights left after pruning,
	especially the number of non-zero weights, $N_{non-zero}$,
		in each weight-fetching group.

This non-uniform distribution can cause a misalignment
	between the activations and weights.
%If an accelerator exploits the sparsity after pruning,
%	the constraint of the accelerator should be considered.
In an accelerator with MWMA PEs,
	the multiplier operands should be fetched simultaneously.
When $N_{par}$ activations and $N_{mul}$ non-zero weights are fetched
	from the internal buffers, it is not guaranteed that every fetched weight
	can find its counterpart in the fetched activations.
Another activation-fetching group may have to be fetched for the process of
	the weights.

To solve the misalignment problem,
	Cambricon-X introduces a padding-zero scheme.
%		where padding-zeros are inserted so that
%			the number of non-zero weights and padding-zeros
%				for each activation-fetching group
%			is always a multiple of $N_{mul}$, 16.
These padding zeros, however, not only waste the internal weight buffer
	but also cause another type of inefficiency.
The total number of padding zeros would be smaller with a larger $N_{par}$.
%A simply estimated number of required padding zeros
%	for AlexNet pruned by \cite{han15}
%	is 0.3M for $N_{par}$=64 and 0.1M for $N_{par}$=256
%	with 0.8M non-zero weights.
This appears one of the reasons why
	Cambricon-X uses a very large $N_{par}$ compared to $N_{mul}$.
For a usual pruning ratio of 75\% in convolutional layers \cite{han15},
	$N_{par}$=64 may be enough for $N_{mul}$=16,
	but Cambricon-X uses $N_{par}$=256.
Due to the large $N_{par}$, the activation selection part,
	the IM block, has very wide, 256-to-1, MUXs.
Because the number of the MUXs is also large, $N_{mul}{\cdot}b$ MUXs per PE,
	these wide MUXs cause a large IM block area,
		occupying more than 30\% of the total chip area.
The large $N_{par}$ also requires a very wide ($N_{par}{\cdot}b$ bit-width)
	internal activation buffer.
This type of wide memory usually induces a larger area than 
	that associated with a square-shaped memory.

Furthermore, the load balance between PEs is also a problem.
A PE processes an activation-fetching group
	for $\lceil N_{non-zero}/N_{mul} \rceil$ cycles.
Owing to the diversity of $N_{non-zero}$,
	the number of cycles will vary as well.
This may cause a problem in certain types of architectures such as Cambricon-X,
	which shares the fetched activations between PEs.
%		to reduce the required bandwidth
%			between the internal buffers and the external memories.
If some PEs complete the process of fetched activations early,
	the PEs must wait for the other PEs to finish.
The load-imbalance problem is also a major concern in accelerators
	with a weight serial structure such as the SWMA and SWSA structures
	\cite{han16_2, han17, kim17}.

There are few studies on pruning with considering the divergent distributions
	of remaining non-zero weights.
The study presented in \cite{boo17} considers
	the distribution of non-zero weights,
	but deals with only the fully-connected layers.
Furthermore, the target was to reduce the width of the ternary weight coding
	without consideration of the accelerator architectures.
A load-balance-aware pruning scheme was proposed in \cite{han17},
%	which prunes the weights so that
%		the total number of nonzero weights processed in a PE
%			is balanced with that in another PE.
	but this pruning scheme focused solely on fully-connected layers, 
	and the accelerator architecture discussion is insufficient, too.
In contrast,
	the proposed scheme, which will be presented in the next section,
	is closely related to accelerator architectures,
	resolving all of the mentioned inefficiency problems.
Furthermore, our scheme focuses on both convolutional layers
	and fully-connected layers.
	
\section{Accelerator-Aware Pruning}
In this section, we propose an accelerator-aware pruning algorithm
	that generates a more regular non-zero weight pattern
	that fits accelerator architectures well.
There can be various architecture consideration points,
	and this paper will concentrate on two parameters,
		the activation- and weight-fetching groups
	and the number of non-zero weights left in each weight group.
These two parameters are closely related to accelerator architectures.
The size of the activation- and weight-fetching group determines
	the internal buffer width,
	and the number of non-zero weights is associated with the required number
		of multipliers and processing cycles.
Previous pruning schemes do not consider these points,
	creating irregular distributions of the non-zero weights
		and the problems mentioned in the previous section.
We will discuss pruning approaches for the architectures
	mentioned in the previous section,
	but the algorithm is not limited to these architectures.

\subsection{Proposed Accelerator-Aware Pruning Scheme}

\begin{figure}[!t]
	\centering
	\includegraphics[scale=0.5]{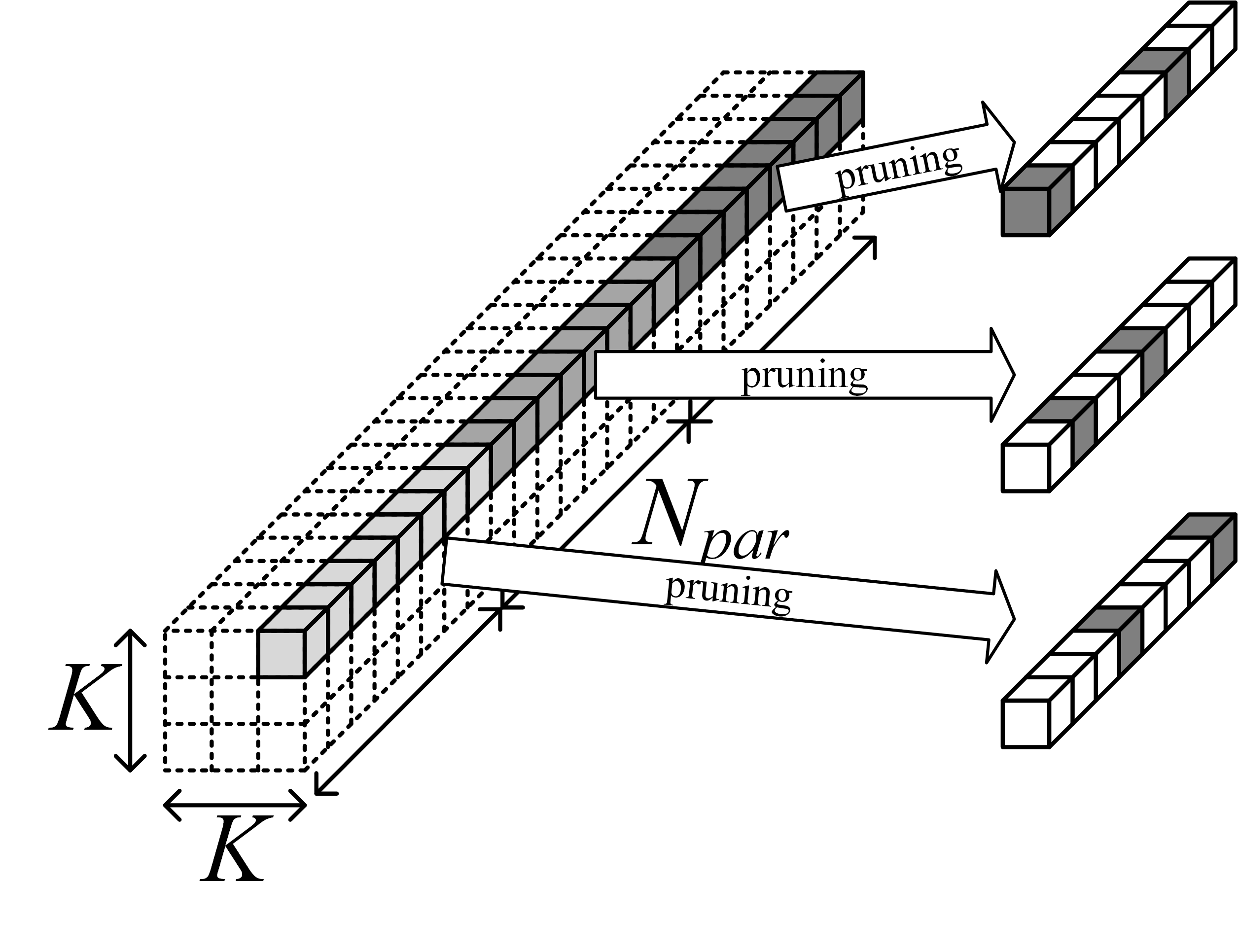}
	\caption{Example result of the proposed pruning scheme
		where $N_{non-zero}$=2 for every weight-fetching group
		along the channel axis ($N_{par}$=8).}
	\label{fig:prune_ex}%
\end{figure}

In the proposed pruning scheme, the weights are pruned
		within the weight-fetching groups
	so that the number of remaining non-zero weights, $N_{non-zero}$,
		is uniform for all of the weight-fetching groups.
The accelerator in \reffig{fig:sparse_ex}, for example,
	simultaneously fetches and processes an activation-fetching group
		consisting of eight activations along the channel axis
		($N_{par}$=8).
$N_{non-zero}$ for the corresponding weight-fetching group is three
	in the figure.
However, previous pruning schemes provide no guarantee for $N_{non-zero}$.
$N_{non-zero}$ can be two or four in another weight-fetching group.
Even zero or the size of a weight-fetching group is possible.
In contrast, the proposed scheme leaves a fixed number of non-zero weights
		for all of the weight-fetching groups,
			as shown in \reffig{fig:prune_ex}.
In the figure, every weight-fetching group has six weights pruned away
	with two non-zero weights remaining ($N_{non-zero}$=2),
		as indicated in white and grey, respectively.

The result of the proposed pruning scheme can be applied
	to the previous sparse accelerators,
	resolving the inefficiency problems mentioned in the previous section.
The misalignment problem in the sparse MWMA and MWSA structures
	will be solved if the number of remaining non-zero weights per group
	is set to be a multiple of $N_{mul}$.
This alignment would make the padding zeros obsolete in Cambricon-X.
Furthermore, the proposed scheme can solve the load-imbalance problem, 
	one of the main concerns in the weight serial structures as well.
Every weight-fetching group has the same number of remaining non-zero weights,
	so the number of weights for a PE to process is naturally balanced
		if every PE processes the same number of weight-fetching groups.
Every sparse architecture mentioned in the previous section
	can benefit from the proposed pruning scheme.

A similar scheme was presented in \cite{boo17},
	but the regularity was used only to reduce the amount of weight storage.
They did not consider the effect on the accelerator architecture.
Furthermore, they only dealt with fully-connected layers.
The load-balance-aware pruning scheme, proposed in \cite{han17},
	adopts a similar concept, but it can be thought of as a special case
	of the proposed scheme.
The load-balance-aware pruning is also limited to fully-connected layers
	and the EIE and ESE architectures.

\subsection{Accelerator Complexity Reduction}
\label{subsec:acc}
\begin{figure}[!t]
	\centering
	\includegraphics[scale=0.75]{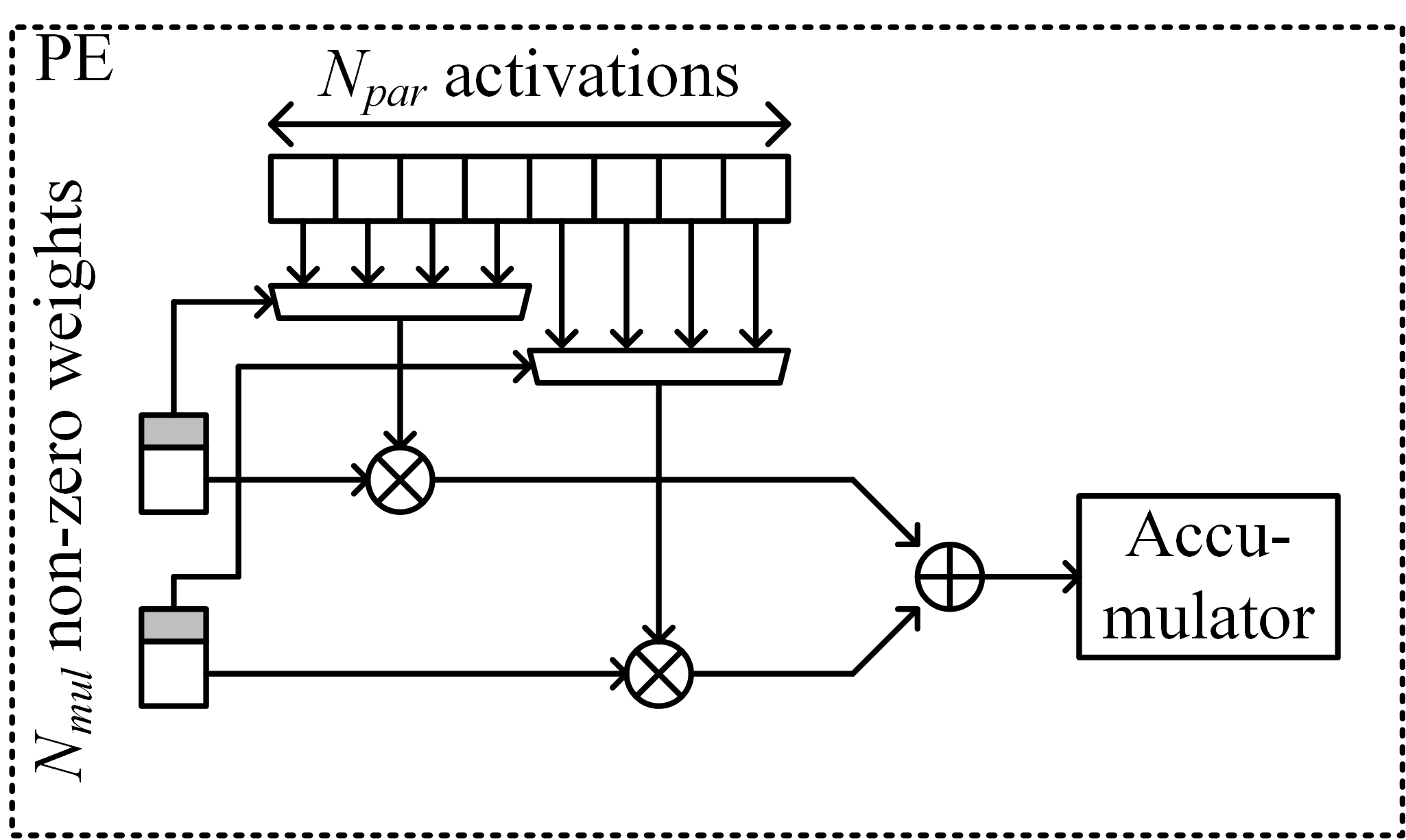}
	\caption{PE structure when $N_{par}$=8 and $N_{group}$=4.}
	\label{fig:small_g}%
\end{figure}

In addition to improving the efficiency of previous sparse architectures,
	the proposed scheme can also be used
		to reduce the degree of accelerator complexity,
		especially with regard to the indexing and activation selection logic.
Cambricon-X has very wide (256${\cdot}b$-bit width) activation buffers
	and very wide (256-to-1) MUXs in the activation selection logic
	to deal with the irregular distribution of non-zero weights
		by the previous pruning schemes,
	as mentioned in Subsection \ref{subsec:prev}.
The input width of the MUXs can be narrowed by the proposed pruning scheme,
	simplifying the activation selection logic.
First, every weight-fetching group is divided evenly into $g$ sub-groups,
		referred to as the pruning groups.
The size of a pruning group, $N_{group}$, will be $N_{par}/g$.
The pruning is then performed so that each pruning group has a constant number
		of weights pruned.
If the number of weights pruned away in a pruning group is $N_{prune}$,
	the pruning ratio will be $N_{prune}/N_{group}$.

Since a weight corresponds to one of the $N_{group}$ activations,
	the width of a MUX for the activation selection
		can be reduced to $N_{group}$.
With the smaller $N_{group}$, the PE structure in \reffig{fig:pe_mwma_sparse}
	can be modified, becoming the structure shown in \reffig{fig:small_g}.
As indicated in the figure, the activation selection logic
	becomes simplified with narrower MUXs.
When $N_{group}$ becomes smaller, however,
	degradation of the network performance can increase
		with the same pruning ratio.
The experiment shows that $N_{group}$=16 with a 75\% pruning ratio
	does not deteriorate the CNN performance.

The proposed pruning scheme can induce a reduction of $N_{par}$, too.
As mentioned in Subsection \ref{subsec:prev}, 
$N_{par}$=256 for Cambricon-X is large compared to $N_{mul}$=16.
$N_{par}$=64 is enough for the common pruning ratio of 75\%
	in the convolutional layers.
A large $N_{par}$ may be chosen to reduce the number of padding zeros.
However, the proposed scheme makes the padding zeros unnecessary,
	implying that $N_{par}$=64 can be used.
In this case, the width of the activation buffers can be reduced to 64,
	enabling more square-like memory components to be used.
Square-like memory components are more area-efficient
	than wide memory components.

Furthermore, the indexing logic can be simplified.
For the indexing of the irregular non-zero weights,
	Deep Compression and EIE use
		relative indexing \cite{han16, han16_2},
	where the number of zero weights between two adjacent non-zero weights
		is stored.
The interval is encoded with four bits, and 
	an interval larger than the encoding bound requires
		filler zero insertion.
A similar indexing scheme is used in Cambricon-X, known as step indexing.
The indexing can be simplified with the proposed pruning scheme,
	where pruning is performed within a pruning group of $N_{group}$.
The small $N_{group}$ enables direct indexing, 
	where an index indicates the position of the non-zero weight
		within the pruning group.
Given that $N_{group}$ is small, the indexing bit width is also small,
		$\lceil \log_2 N_{group} \rceil$ bits,
	even with direct indexing.
Direct indexing is much simpler than 
	relative indexing in EIE or step indexing in Cambricon-X
	and does not require filler zeros,
		removing the waste of the weight storage.

%\subsection{Pruning Ratio and Pruning Weight Selection}
%The proposed pruning scheme can set the pruning ratio target directly,
%	as apposed to the previous schemes.
%In \cite{han15}, weights with a magnitude less than a threshold
%	are pruned away.
%The threshold is determined according to the sensitivity and the
%	weight value standard deviation of each layer.
%With this scheme, the pruning ratio cannot be predicted easily.
%The pruning ratio is determined after pruning.
%If the target ratio is not reached, the pruning is tried again
%	with a smaller threshold.
%The process is iterated until the target pruning ratio is reached.
%In the proposed pruning, however, the target is the number of pruned weights
%	per pruning group.
%In other words, the pruning ratio is directly focused.
%If the number of weights pruned away in a pruning group is $N_{prune}$,
%	the pruning ratio will be $N_{prune}/N_{group}$.
%
%Various methods can be used to select the weights to be pruned.
%The simplest one is to select the weights by weight magnitude.
%In a more complex method, the effect of weights is estimated
%	and the weights with the a smaller effect are removed earlier.
%The proposed pruning scheme can be used with any selection method,
%	but, in this paper, the simplest method will be used.
%The weights with the least magnitude are pruned first in a pruning group.

\subsection{Incremental Pruning}
In the proposed scheme, $N_{prune}$ weights are pruned in a pruning group.
The pruning can be processed in a few different ways.
At one extreme, the target number of weights with the least magnitude
	are pruned at the same time in each group,
	and the pruned network is retrained.
This scheme is called one-time pruning in this paper.
At the other extreme, pruning begins with only one weight
	with the least magnitude in each group.
Subsequently, a period of retraining is undertaken, followed by the pruning of
	one more weight with the least retrained magnitude.
Retraining is performed again.
The one weight pruning and retraining processes are iterated
	until the target number is reached.
In the middle of the two extreme methods,
	we can set the initial pruning number and the increment number.
This scheme is referred to here as incremental pruning.

\begin{figure}[!t]
	\centering
	\includegraphics[scale=0.7]{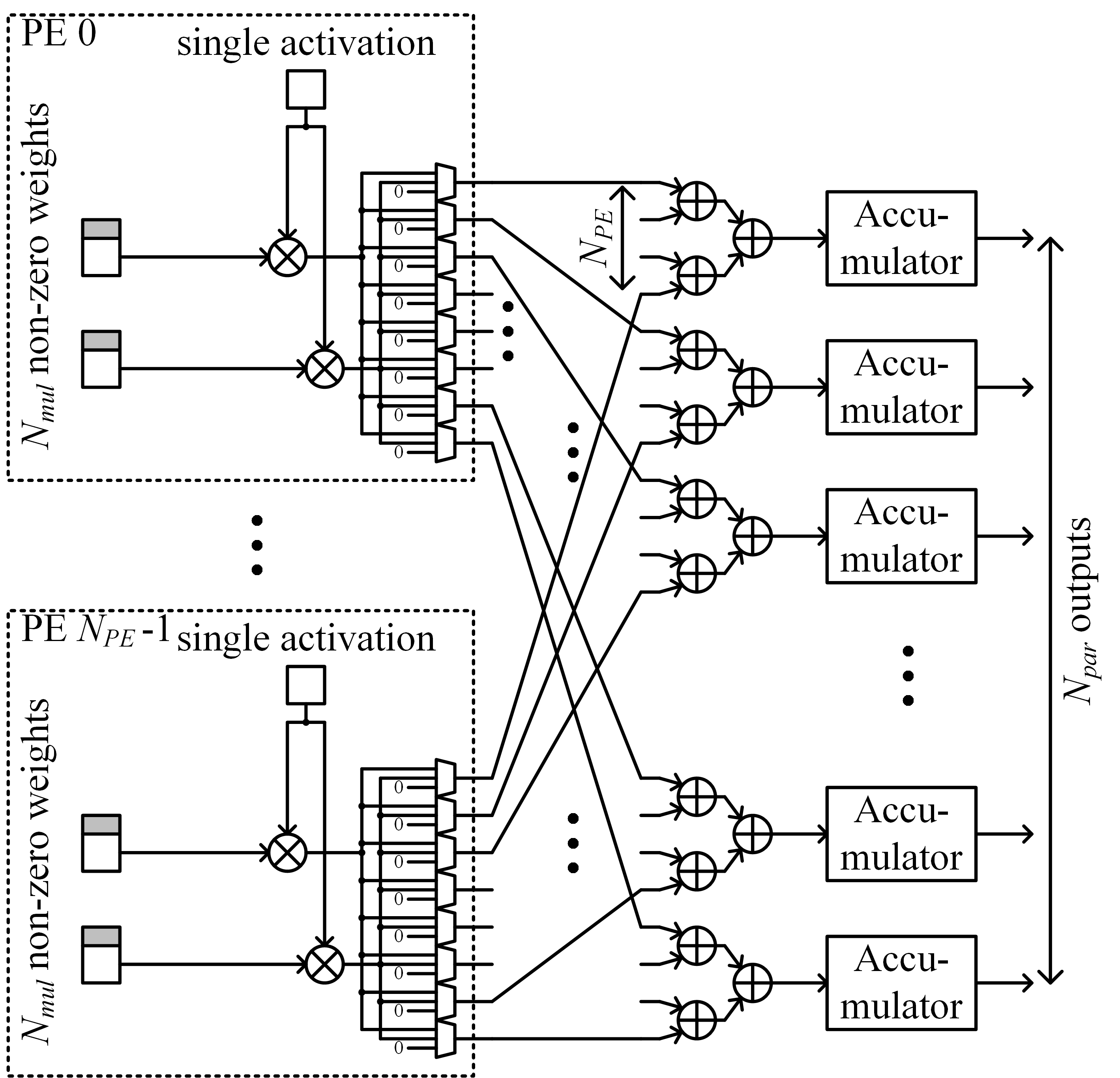}%
	\caption{Sparse MWSA PE structure.}
	\label{fig:pe_mwsa_sparse}%
\end{figure}

Obviously, incremental pruning would be better than or equal to
	the one-time pruning method.
However, incremental pruning requires a long retraining time.
	Accordingly, in this paper, it will be applied when one-time pruning
		is not sufficient.

\subsection{Pruning for Various Architectures}

In the proposed pruning scheme, pruning can be applied to CNNs
	for various accelerator architectures.
In the previous subsections, pruning along the channel axis is shown
	for architectures such as Cambricon-X.
As an example of pruning for another type of accelerator architecture,
	the proposed pruning scheme can be adjusted to 
		a weight-sparse version of Cnvlutin.
Cnvlutin adopts the MWSA structure, and the weights are fetched
	along the filter axis.
In this architecture, the dense MWSA structure can be modified
	to the sparse MWSA structure shown in \reffig{fig:pe_mwsa_sparse}.
For the sparse architecture, the proposed pruning scheme can be applied
	to CNNs with pruning groups set to the weights along the filter axis.
Because Cnvlutin already exploits activation sparsity,
	the structure in \reffig{fig:pe_mwsa_sparse} can exploit
		both activation sparsity and weight sparsity.

The architectures in \cite{qiu16, motamedi16}
	adopt sparse MWMA structure PEs, where the activations and weights
		are fetched and processed along the spatial axis.
For these architectures, the proposed pruning scheme can be applied
	with pruning groups established along the spatial axis.

%The SCNN architecture presented in \cite{parashar17}
%	exploits both of weight sparsity and activation sparsity.
%For the performance improvement, a special structure of cartesian product
%	and a complex output distribution circuit are required.
%Since this accelerator proceses parallely weights in the spatial axis,
%	the pruning group can be constituted with the $K \times K$ weights
%		along the spatial asixs, too.

\subsection{Application to Fully-Connected Layers}
The proposed pruning scheme can be applied to fully-connected layers as well.
In a fully-connected layer, the weights can be arranged in a matrix format.
We can define two axes: the column axis and the row axis.
Along the row axis, the weights are multiplied with different activations,
	while along the column axis, the weights are multiplied with an activation.
An MWMA-structured PE processes the weights along the row axis simultaneously,
	and an MWSA-structured PE processes the weights along the column axis.

\begin{figure}[!t]
	\centering
	\includegraphics[scale=0.5]{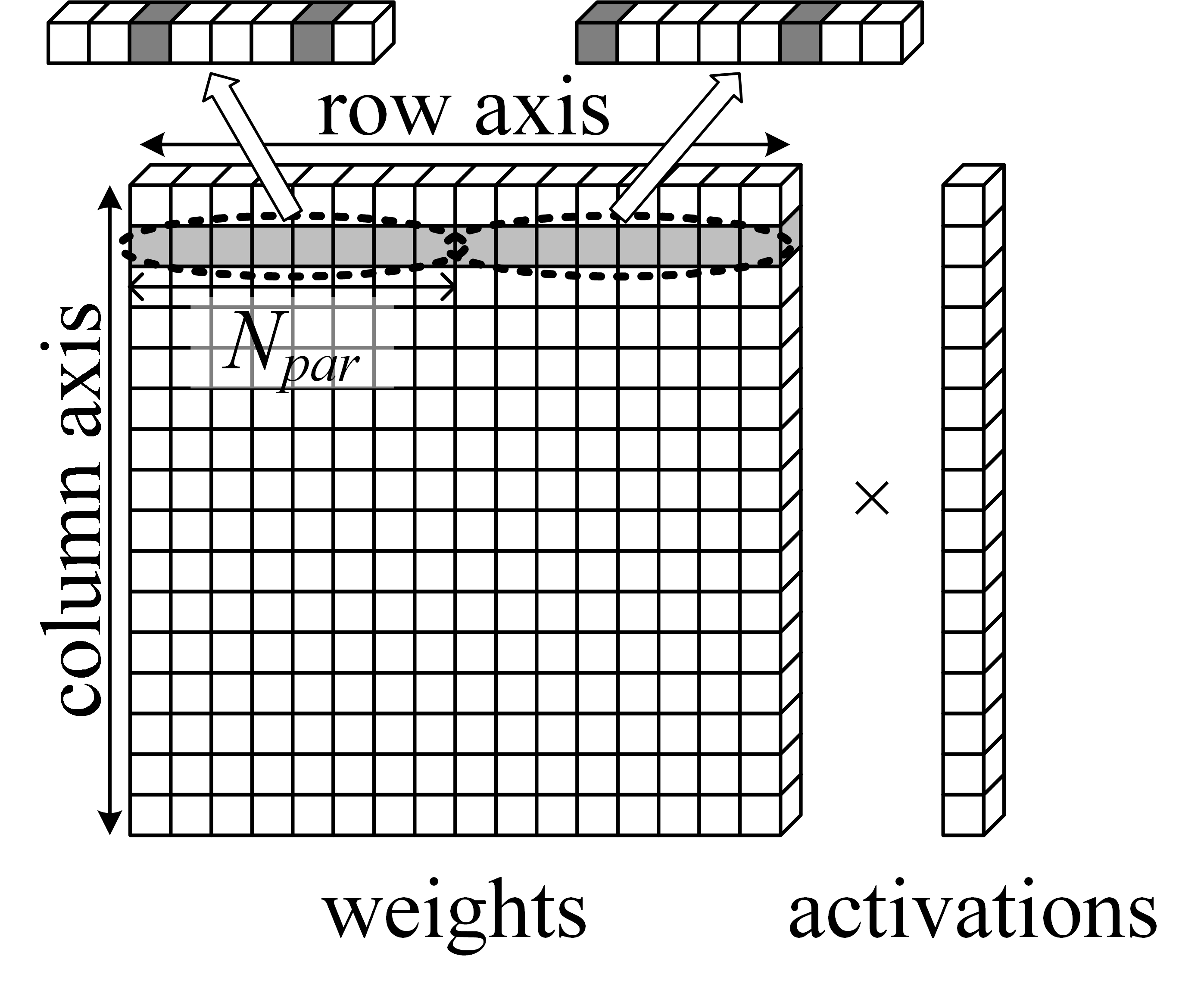}
	\caption{Example result of the proposed pruning scheme
		with $N_{group}$=8 and $N_{prune}$=6 on a fully-connected layer.}
	\label{fig:prune_fc_ex}%
\end{figure}

According to the PE structures,
	accelerator-aware pruning is applied following different axes.
If an MWMA-structured PE is used, the weights are grouped along the row axis
	and pruned so that each group has
		a fixed number of non-zero weights remaining.
If an MWSA-structured PE is used, the weights are grouped and pruned
	along the column axis.
An example of the proposed pruning scheme along the row axis
	is illustrated in \reffig{fig:prune_fc_ex}.
In the figure, $N_{par}$=$N_{group}$=8 and $N_{prune}$=6.

%\subsection{Effect on Weight Serial Structures}
%Some architectures adopt PEs of the weight serial structure,
%	where a weight is fetched and multiplied
%		with multiple activations (SWMA) or a single activation (SWSA).
%These architectures can accept easily irregular pattern
%	of unstructured weight pruning.
%Therefore, the SWSA structure is adopted
%		in EIE \cite{han16_2} and ESE \cite{han17},
%	which are designed for the fully-connected layers
%		pruned by the unstructured pruning of \cite{han15} and \cite{han16}.
%
%However, unstructured pruning causes a load-balance problem,
%	where PEs take different time to process.
%To solve this problem, a laod-balance-aware pruning has been proposed
%	in \cite{han17},
%		where the weights are pruned so that the toatl number of weights
%			that each PE processes is equal or similar
%		to that of the the others.
%The proposed accelerator-aware pruning constrains
%	the number of non-zero weights for each pruning group.
%Therefore, the load-balancing is accomplished naturally.
%
\section{Experimental Results}
To show that the proposed pruning scheme can preserve
	the performance of CNNs well even with the given constraint,
	the top-5 accuracy for the ImageNet 2012 validation data set
	\cite{russakovsky15}
		was measured.
The retraining was performed by Caffe \cite{jia14} in one of three modes.
In Retraining 1, after one-time pruning, retraining was performed
		with a learning rate of 5$\cdot$10$^{-4}$ for 12 epochs.
If the original accuracy was not recovered,
	8-epoch retraining was performed additionally
		with a learning rate of 10$^{-4}$.
If Retraining 1 was not enough, Retraining 2 was applied,
	where the learning rate begins at 5$\cdot$10$^{-4}$ and decreases to 
	10$^{-4}$, 10$^{-5}$, and 10$^{-6}$
		when the validation accuracy becomes saturated.
In Retraining 3, Retraining 2 is applied with incremental pruning.
When the validation accuracy is saturated at the learning rate of 10$^{-6}$,
	$N_{prune}$ is increased and
	the retraining resumes with a learning rate of 5$\cdot$10$^{-4}$.
The retraining mini-batch size was set to 256.
The network models were obtained publicly
		\cite{caffe, vgg, resnet, squeezenet, mobilenet_caffe}.
		%, shufflenet_caffe_model}.
Various methods can be used to select the weights to be pruned.
Any selection method is applicable,
	but the simplest method was used in the experiments.
The weights with the least magnitude are pruned first in a pruning group.

The pruning of convolutional layers will be discussed first
	because convolutional layers account for most of the computations.
The proposed pruning scheme does not change the channel number of 
	output feature maps
	and can therefore be applied to the residual blocks of ResNet
		including multiple branches without difficulty.
The bias values are not pruned.
Our pruning scheme was then applied to fully-connected layers
	with the pruned convolutional layers.
It will be shown that the proposed pruning method is also applicable
	to compact networks.
%networks that are already slimmed
%		by the previous channel pruning schemes, too.
The last two subsections will analyze the effects of the proposed scheme
	on the accelerator performance and complexity.

\subsection{Convolutional Layer Pruning}

\begin{table}[!t]
	\processtable{ImageNet validation accuracy(\%) right after convolutional layer pruning}
	{\label{tab:prune1}}
	{\begin{tabular}[!t]{r r | r r r r}
	\hline
	 $N_{group}$ & $N^{conv}_{prune}$ & AlexNet & VGG16 & ResNet-50 & ResNet-152 \\
	\hline
	\hline
	- & - & 79.81 & 88.44 & 91.14 & 92.20 \\
	\hline
	8 & 1 & 79.68 & 88.31 & 90.94 & 92.12 \\
	8 & 2 & 79.17 & 88.02 & 90.30 & 91.69 \\
	8 & 3 & 76.89 & 85.52 & 88.27 & 90.72 \\
	8 & 4 & 65.02 & 70.20 & 78.79 & 86.37 \\
	8 & 5 & 32.19 & 12.23 & 17.21 & 46.58 \\
	\hline
	16 & 1 & 79.83 & 88.47 & 91.16 & 92.22 \\
	16 & 2 & 79.76 & 88.36 & 91.08 & 92.18 \\
	16 & 4 & 79.35 & 88.19 & 90.80 & 92.07 \\
	16 & 6 & 78.00 & 86.64 & 88.88 & 91.36 \\
	16 & 8 & 69.99 & 75.34 & 80.74 & 87.90 \\
	16 & 9 & 60.84 & 55.15 & 59.39 & 80.36 \\
	16 & 10 & 38.61 & 20.12 & 32.42 & 65.67 \\
	\hline
\end{tabular}
} {}
\end{table}

\begin{table}[!t]
	\processtable{ImageNet validation accuracy(\%) after convolutional layer pruning and retraining}
	{\label{tab:retrain1}}
	{
	\begin{tabular}[!t]{r r | r r r r}
	\hline
	 $N_{group}$ & $N^{conv}_{prune}$ & AlexNet & VGG16 & ResNet-50 & ResNet-152 \\
	\hline
	\hline
	- & - & 79.81 & 88.44 & 91.14 & 92.20 \\
	\hline
	8 & 4 & 80.45 & 90.71 & 91.95 & 93.03 \\
	8 & 5 &\textsuperscript{*}80.58 & 90.45 & 91.68 & 92.79 \\
	8 & 6 &\textsuperscript{*}80.42 & 89.91 & 91.14 & 92.33 \\
	8 & 7 &\textsuperscript{*}79.47 & 88.48 & 88.86 & 91.00 \\
	\hline
	16 & 8 & 80.46 & 90.73 & 91.96 & 93.02 \\
	16 & 9 & 80.38 & 90.80 & 91.80 & 92.89 \\
	16 & 10 &\textsuperscript{*}80.62 & 90.54 & 91.86 & 92.75 \\
	16 & 11 &\textsuperscript{*}80.74 & 90.38 & 91.55 & 92.55 \\
	16 & 12 &\textsuperscript{*}80.50 & 90.22 & 91.35 & 92.48 \\
	16 & 13 &\textsuperscript{*}80.22 & 89.65 & 90.88 & 92.20 \\
	\hline
	 \multicolumn{2}{c|}{Unstructured (81.25\%)} &\textsuperscript{*}80.34 & 90.06 & 91.17 & 92.37 \\
	\hline
\end{tabular}

	} {}
\end{table}

\reftab{tab:prune1} shows the accuracy results right after the pruning
	of the convolutional layers.
The weights are grouped along the channel axis,
	and the first convolutional layer is not pruned
		as it has a much smaller number of weights and operations
			than the other layers.
The table shows that 30\% (3/8 or 6/16) pruning already begins
	to degrade the accuracy.
However, the degradation can be recovered with retraining
	as shown in \reftab{tab:retrain1}.
Retraining 1 and 3 are applied;
	the results of Retraining 3 are denoted by asterisks.

In the table, with up to 75\% (6/8 or 12/16) pruning, the validation accuracy 
	could be recovered to the baseline accuracy with Retraining 1
		in very deep networks, including VGG16, ResNet-50, and ResNet-152.
The result of VGG16 matches that of the unstructured pruning algorithm
	in \cite{han15},
	where the pruning ratios of the convolutional layers
		are more or less than 75\% in VGG16.
The experiment shows that a similar pruning ratio can be reached 
	with the proposed pruning scheme considering the accelerator constraint.
It can also be seen that a 75\% pruning ratio does not degrade the accuracy
	in the residual networks, ResNet-50 and ResNet-152.
The networks have more complicated structures,
	such as a residual path and 1$\times$1 convolution.
The results show that the proposed scheme can be applied
	to recent state-of-the-art CNNs.
With some pruning ratios, the accuracy is improved,
	which has been observed in other pruning papers, too \cite{han15, mao17}.
This improvement appears to be caused
	by a kind of regularization \cite{mao17}.

In a relatively shallow network such as AlexNet,
	it was more difficult to recover the accuracy.
With the more advanced effort of Retraining 3, however,
	the original accuracy level can be recovered with pruning ratios comparable
		to those in \cite{han15}.
%AlexNet has convolutional layers with large spatial dimension
%	and less channel dimension.
In AlexNet, Retraining 3 begins with ($N_{prune},N_{group}$)=(5,8) or (10,16),
	and $N_{prune}$ is increased by one.
While the pruning ratio of the AlexNet convolutional layers was around 65\%
	with the unstructured pruning in \cite{han15}, 
	the proposed scheme can reach a pruning ratio of 81.25\%
		after approximately 300 epochs of retraining.

The last row of the table shows the results of an unstructured pruning scheme,
	which prune 81.25\% of weights with the least magnitude
	in each convolutional layer.
The unstructured pruning scheme shows a little better accuracies,
	but the differences are negligible.

\subsection{Fully-Connected Layer Pruning}

\begin{table}[!t]
	\processtable{ImageNet validation accuracy(\%) after fully-connected layer pruning and retraining with $N_{group}$=16 and $N^{conv}_{prune}$=12}
	{\label{tab:retrain_fc}}
	{
	\begin{tabular}[!t]{r r| r r r r}
	\hline
	 $N^{fc1,2}_{prune}$ & $N^{fc3}_{prune}$ & AlexNet & VGG16 & ResNet-50 & ResNet-152 \\
	\hline
	\hline
	- & - & 79.81 & 88.44 & 91.14 & 92.20 \\
	\hline
	12 & 12 &\textsuperscript{*}80.24 & 89.84 & 91.24 & 92.54 \\
	13 & 12 &\textsuperscript{*}80.29 & 89.56 & - & - \\
	14 & 12 &\textsuperscript{*}80.01 & 89.19 & - & - \\
	15 & 12 &\textsuperscript{*}79.47 & 88.95 & - & - \\
	\hline
\end{tabular}

	} {}
\end{table}

After the pruning of the convolutional layers,
	the fully-connected layers were pruned along the row axis.
In the fully-connected layer pruning step, 
	we attempted to prune more weights 
	than were pruned in the convolutional layers
	because, generally, more weights can be pruned
		in fully-connected layers \cite{han15}.
%So, this policy will give less number of remaining weights.
%With less number of weights in a group than the number of multipliers,
%	some multipliers will perform no valid operation.
%However, the effective throughput will not be degraded
%	because less number of operations are processed.
The size of a pruning group, $N_{group}$, was equally set
	for the convolutional layers and fully-connected layers.

The retrained accuracy is shown in \reftab{tab:retrain_fc}.
In the table, $N_{prune}^{conv}$, $N_{prune}^{fc1,2}$, and $N_{prune}^{fc3}$
	are the number of pruned weights
	in the convolutional layers, the first and second fully-connected layers,
	and the last fully-connected layer, respectively.
ResNet-50 and ResNet-152 have one fully-connected layer;
	hence, $N_{prune}^{fc1,2}$ was ignored in these networks.
Retraining 1 was applied to all of the networks except for AlexNet,	
	to which Retraining 3 was applied.
In Retraining 3, $N_{prune}^{fc1,2}$ was increased by one from 12.

The table shows that the proposed pruning scheme can reach a pruning ratio 
	similar to that of the previous pruning scheme
		in the fully-connected layers.
In \cite{han15}, 90--96\% of the weights were pruned in the first and
	the second fully-connected layers of AlexNet and VGG16,
	and as were 75--77\% of the weights in the last layer.
In all of the presented networks, the proposed pruning scheme
	could prune 75\% of the weights ($N_{prune}^{fc3}$=12)
		in the last fully-connected layers.
For the first and second fully-connected layers of VGG16,
	pruning with $N_{prune}^{fc1,2}$=15 (93.75\% pruning)
		did not degrade the accuracy.
In AlexNet, $N_{prune}^{fc1,2}$=15 showed an accuracy of 79.47\%,
		which is slightly worse
			than the accuracy of the pruned AlexNet in \cite{han15}, 79.68\%.
Because $N_{prune}^{conv}$=12 in the convolutional layers is larger
	than 62--65\% in \cite{han15}, the pruning results are quite comparable.

\subsection{Pruning Along Other Axes}

\begin{table}[!t]
%\begin{threeparttable}
	\processtable{ImageNet validation accuracy(\%) after pruning and retraining
		along other axes} 
	{\label{tab:retrain_axes}}
	{\begin{tabular}[!t]{l r r | r r}
	\hline
	 Axis & $N_{group}$ & $N^{conv}_{prune}$ & VGG16 & ResNet-50 \\
	\hline
	\hline
	 - & - & 0 & 88.44 & 91.14 \\
	\hline
	 Filter &16 & 10 & 90.63 & 91.71 \\
	 Filter &16 & 11 & 90.46 & 91.54 \\
	 Filter &16 & 12 & 90.12 & 91.31 \\
	 Filter\textsuperscript{\#} &16 & 12 & 89.49 & 90.85 \\
	\hline
	 Spatial &9 & 5 & 90.32 & 91.85 \\
	 Spatial &9 & 6 & 89.69 & 91.49 \\
	 Spatial &9 & 7 & 88.86 & 91.31 \\
	\hline
\end{tabular}
} {}
%	\begin{tablenotes}
%	\item [a] no pruning in fully-connected layers
%	\item [b] $N^{fc1,2}_{prune}$=15 and $N^{fc3}_{prune}$=12
%	\end{tablenotes}
%\end{threeparttable}
\end{table}

The proposed pruning scheme can be applied along other axes.
The accuracy results after pruning and retraining
	along the filter and the spatial axis
	are presented in \reftab{tab:retrain_axes}.
At the third to fifth rows in this table,
	the convolutional layers are pruned along the filter axis,
	and at the next row, denoted by '\#', the fully-connected layers are 
	further pruned with $N_{prune}^{fc1,2}$=15 and $N_{prune}^{fc3}$=12.
At the last three rows, the convolutional layers are pruned
	along the spatial axis.
These results show that the accuracy is not degraded
	by pruning along other axes,
	meaning that the proposed accelerator-aware pruning scheme can be used
	for various accelerator architectures.

\subsection{Compact Network Pruning}

\begin{table}[!t]
	\processtable{ImageNet validation accuracy(\%) after pruning and retraining
		of compact networks}
	{\label{tab:retrain_light}}
	{
	\begin{tabular}[!t]{r r | r r}
	\hline
	 $N_{group}$ & $N^{conv}_{prune}$ & SqueezeNet v1.0 & MobileNetV1-224 1.0\\
	\hline
	\hline
	- & - & 80.39 & 89.24 \\
	\hline
	16 & 8 &\textsuperscript{*}80.93 & 89.97 \\
	16 & 9 &\textsuperscript{*}80.65 & 89.67 \\
	16 & 10 &\textsuperscript{*}80.29 &\textsuperscript{*}89.79 \\
	16 & 11 &\textsuperscript{*}79.86 &\textsuperscript{*}89.50 \\
	16 & 12 &\textsuperscript{*}78.80 &\textsuperscript{*}89.06 \\
	\hline
\end{tabular}

	} {}
\end{table}

Some compact networks have been proposed recently
	\cite{iandola16, howard17, zhang18}.
The compact networks are less redundant,
	meaning that pruning may be more harmful.
	%so it may be harder to recover the accuracy degradation after pruning.
\reftab{tab:retrain_light}, however,
	shows that the original accuracy can be recovered
		with the accelerator-aware pruning along the channel axis.
In SqueezeNet, the proposed scheme can reach a pruning ratio comparable
	to 66.3\%, the pruning ratio of an unstructured pruning scheme
	in \cite{iandola16}.
%75\% pruning degrades the accuracy slightly.
With $N_{prune}^{conv}$=10 or 11,
	the accuracy degradation of the proposed pruning scheme
	is around 0.5 percent points.
In MobileNet \cite{howard17}, the accuracy is recovered more easily.
In this case, 75\% pruning shows still good accuracy.
MobileNet consists of depthwise convolutional layers
	and pointwise convolutional layers,
	and only the latter ones were pruned because they account for
		most of the operations and weight storage.

\subsection{Slimmed Network Pruning}
\begin{table}[!t]
	\processtable{ImageNet validation accuracy(\%) after pruning and retraining
		of slimmed networks in \cite{he17}}
	{\label{tab:retrain_slimmed}}
	{
	\begin{tabular}[!t]{r r | r r r}
	\hline
	 $N_{group}$ & $N^{conv}_{prune}$ & VGG16-4X & VGG16-5X & ResNet-50 CP \\
	\hline
	\hline
	- & - & 89.06 & 86.98 & 89.64 \\
	\hline
	16 & 8 & 89.53 & 88.25 & 91.44 \\
	16 & 10 & 88.58 & 87.91 & 91.28 \\
	16 & 12 &\textsuperscript{*}87.05 & 87.36 & 90.30 \\
	\hline
\end{tabular}

	} {}
\end{table}

Some previous works pruned convolutional layers
	in a channel-wise or filter-wise approach \cite{wen16, li17, he17}.
The resulting networks are slimmer networks with fewer channels
	in the layers.
%These schemes have an advantage
%	that the implementation method of the dense networks
%		can be used without modification.
Slimmed networks can also be pruned 
	by the proposed pruning scheme.
For this experiment, the networks slimmed in \cite{he17} were used because
	their models are publicly available \cite{heyihui}.
In \cite{he17}, the weight amount of the networks was also
	reduced by other methods such as decomposition.
%The slimmed networks have 50\% to 70\% of weights in the convolutional layers
%	compared to the original networks.
The proposed scheme was applied with $N_{group}$=16.
In some layers of the slimmed networks, $C$ is not a multiple of $N_{group}$.
In such a case, it is assumed that zero weights are added
	to make $C$ a multiple of $N_{group}$.

The accuracy results are shown in \reftab{tab:retrain_slimmed}.
The table presents that the proposed pruning scheme
	prunes weights fairly well even in the already slimmed networks.
In this case, 50\% pruning  of the convolutional layers ($N_{prune}^{conv}$=8)
	does not degrade the accuracy of the slimmed networks.
%VGG16 4X: from 14710464 --> 5319133
%Since the slimmed networks have already around half weights pruned away,
%	the additional 50\% pruning reaches around 75\% pruning totally.
%Since the slimmed networks have already less weights than the original
%	VGG16 or ResNet-50 networks,
%	the 50\% additional pruning would give  
It was also noted that 75\% pruning of $N_{prune}^{conv}$=12 does not degrade
	the accuracy of VGG16-5X or ResNet-50 CP.

\subsection{Accelerator Performance}
\begin{table}[!t]
	\processtable{Accelerator performance estimation}
	{\label{tab:proc}}
	{
\begin{tabular}{ l l | r r r r}
	\hline
	Architecture&	& Han15\cite{han15} & Han15\cite{han15}\textsuperscript{\#} & AAP \\ %& AAP (8/16) 
	\hline
	\hline
	Cambricon-X	& $N_{nonzero}^{total}$		& 1098K	& 751K	& 751K \\ %& 1502K	
	Conv2--5	& $N_{padding}$			& 160K	& 171K	& 38K \\ %& 77K	
				& $N_{MAC}$					& 283M	& 191M	& 191M \\ %& 383M	
				& $N_{cycle}$				& 1946K	& 1540K	& 857K \\ %& 1714K	
				%& Utilization (\%)			& 56.76	& 48.56	& 87.24 \\ %& 87.24	
				& Utilization				& 57\%	& 49\%	& 87\% \\ %& 87.24	
	\hline
	EIE			& $N_{nonzero}^{total}$		& 5925K	& 4456K	& 4456K \\ %& 1502K	
	FC1--3		& $N_{cycle}$				& 187K	& 154K	& 70K 	\\ %& 1714K	
				%& Utilization (\%)			& 47.78	& 43.02	& 99.63 \\ %& 87.24	
				& Utilization 				& 49\%	& 45\%	& 100\% \\ %& 87.24	
	\hline
\end{tabular}
	} {}
\end{table}

A neural network pruned by the proposed scheme
	can be processed more efficiently in accelerators.
The upper part of \reftab{tab:proc} compares the efficiency of Cambricon-X
	when it runs the second to fifth convolutional layers of AlexNet
		pruned by the unstructured pruning in \cite{han15}
		and the proposed accelerator-aware pruning.
In the table, $N_{nonzero}^{total}$, $N_{padding}$, $N_{MAC}$, and
	$N_{cycle}$ are the number of total non-zero weights after each pruning,
		the required number of padding zeros for the alignment,
		the number of multiplication and accumulation (MAC) operations
			with non-zero weights,
		and the number of estimated processing cycles, respectively.
The utilization is the ratio of time when the multipliers operate
	with valid operands, calculated as follows:
\begin{equation}
	\text{Utilization} = \frac{N_{MAC}}
			{N_{cycle} \times N_{mul} \times N_{PE}}.
\end{equation}

With the conventional pruning scheme,
	Cambricon-X executes the pruned network with some inefficiency,
	as shown at the third column.
The amount of the required padding zeros is approximately
		14.6\% of the non-zero weight amount,
	which indicates a waste of the internal buffer and the multipliers.
With the additional problem of the load imbalance between PEs,
	the utilization of the multipliers is only 57\%.

With the proposed scheme shown in the last column,
	the resource waste can be greatly reduced.
For the column, AlexNet is pruned by the proposed scheme
	with $N_{prune}^{conv}$=12 and $N_{group}$=16.
The amount of the padding zeros is merely 5\% of the non-zero weight amount.
The padding zeros are required at the second convolutional layer
	because this layer does not have enough input channels.
With 48 input channels, only twelve non-zero weights remain
	for an activation-fetching group,
	which is less than $N_{mul}$.
The small amount of padding zeros and the naturally load-balanced PEs
	lead to high utilization, at 87.24\%,
	and the processing cycles are reduced by 56\%.

Since the pruning ratio of \cite{han15} is less than 75\%,
	we compared an additional case at the fourth column.
For this column, each layer of the network at the third column 
	is additionally pruned to reach 75\% pruning ratio.
With the additional pruning, the numbers of non-zero weights and MACs
		are reduced,
	but a similar number of padding zeros is still required.
The utilization is deteriorated further.
Compared to this case, the proposed pruning scheme can reduce the number of
	processing cycles by 44\%.

The lower part of \reftab{tab:proc} compares the number of processing cycles
	of the EIE architecture.
Because the EIE architecture only deals with the fully-connected layers,
	the table compares the cycles processing the fully-connected layers
		of AlexNet.
In this part, $N_{MAC}$ and $N_{zero-pad}$ are not described
	because $N_{MAC}$ is equal to $N_{nonzero}^{total}$
		in fully-connected layers
	and EIE does not require padding zeros.
In EIE, $N_{PE}$ is 64, and each PE has a multiplier.
According to the EIE process, the accelerator-aware pruning is applied
	along the column axis with $N_{group}$=16, $N_{prune}^{fc1,2}$=15,
	and $N_{prune}^{fc3}$=12.
As shown in the table, the load-imbalance problem degrades the utilization
	to 49\%.
The proposed pruning scheme, shown at the fifth column,
	improves the utilization to 100\%,
	reducing the number of processing cycles by around 63\%.
For a fair comparison, the network of the third column is pruned additionally,
	and the result is shown at the fourth column.
With the utilization under 50\%,
	the additionally pruned network still requires more than twice 
		as many processing cycles as that of the fifth column.

\subsection{Accelerator Logic Complexity}
\begin{table*}[!t]
	\processtable{Accelerator Synthesis Results}
	{\label{tab:synth}}
	{
\begin{tabular}{ l r r r| r r r r r r | r r}
	\hline
	Accelerators			& $N_{par}$ & $N_{group}$ & $N_{mul}$ 
			& \multicolumn{6}{c|}{Area (mm$^2$)}    & Delay & Power \\
	(Measurement Condition) & & &			
			& NBin & NBout & SB & IM & 16 PEs & Total & (ns)  & (W)\\
	\hline
	\hline
	Cambricon-X (P\&R)			& & &
			& 0.55 & 0.55 & 1.05 & 1.98 & 1.78 & 6.38 & 1.00 & 0.95 \\
	Re-implemented (Synthesis)	& 256 & 256 & 16
			& 0.32 & 0.32 & 0.43 & 2.39 & 1.46 & 4.91 & 1.02 & 2.56 \\
	Reduced (Synthesis)			& 64 & 8 & 16
			& 0.11 & 0.11 & 0.43 & 0.17 & 1.19 & 2.01 & 1.05 & 1.24 \\
	Reduced (Synthesis)  		& 64 & 16 & 16
			& 0.11 & 0.11 & 0.43 & 0.27 & 1.41 & 2.34 & 1.02 & 1.49 \\
	Reduced (Synthesis)  		& 64 & 32 & 16
			& 0.11 & 0.11 & 0.43 & 0.44 & 1.44 & 2.53 & 1.02 & 1.58 \\
	Cambricon-X Reduced (Estimation)	& 64 & 16 & 16
			& 0.19 & 0.19 & 1.05 & 0.22 & 1.72 & 3.84 & 1.00 & 0.56  \\
	\hline
	Reduced (Synthesis)  		& 128 & 16 & 16
			& 0.18 & 0.18 & 0.43 & 0.28 & 1.41 & 2.47 & 1.02 & 1.58 \\
	Reduced (Synthesis)  		& 64  & 16 & 8
			& 0.11 & 0.11 & 0.29 & 0.16 & 0.62 & 1.29 & 1.02 & 0.71 \\
	\hline
\end{tabular}
	} {}
\end{table*}

As mentioned in Subsection \ref{subsec:acc},
	the proposed pruning scheme can reduce the accelerator complexity.
In this subsection, attempts are made to reduce the area of Cambricon-X.
The second row in \reftab{tab:synth} presents the area,
	delay time, and power consumption of Cambricon-X.
The power values were measured with dynamic simulation.
In the table, NBin, NBout, and SB are the input activation buffer,
	the output activation buffer, and the weight buffer, respectively.
IM is the indexing and activation selection unit
	including the activation selection MUXs.

Because the RTL code of Cambricon-X is not publicly available,
	we re-implemented the blocks in the table for the area comparison.
The other blocks in Cambricon-X are not affected by the proposed scheme.
Hardware parameters are inferred from the context of the paper.
The synthesis results of the re-implemented case are presented at the third row.
The synthesis was performed by Synopsys DesignCompiler
	with the Global Foundry 65nm process library.
Since the results of Cambricon-X are obtained
	after the placement and routing (P\&R) process with a different library,
	it is difficult to compare the values of the second and third rows directly.
Therefore, the expected reduction effect on Cambricon-X will be estimated
	from the reduction in the re-implemented version,
	which is shown at the following rows.
The number of PEs, $N_{PE}$, is assumed to be 16.
%$N_{par}$ is 256 for Cambricon-X and 64 for the other cases.
%$N_{mul}$ and $N_{PE}$ are 16.

The fourth to sixth rows present the synthesis results
	of reduced implementation
	assuming the application of the proposed pruning scheme with various
		$N_{group}$ values.
Compared to the results of the re-implemented Cambricon-X at the third row,
	the area is reduced greatly, especially that used by NBin, NBout and IM.
The area of NBin and NBout is reduced due to
	the memory width reduction from 256$\times$16 to 64$\times$16.
Although the memory capacity remains unchanged,
	the more square-shaped memory configuration results in a smaller area.
The area reduction of the IM block is more astonishing,
	with a decrease of 82--93\%.
The wide MUXs, 256-to-1 MUXs, in the IM block of Cambricon-X
	are substituted with narrow $N_{group}$-to-1 MUXs.
Because the area of a MUX logic is proportional to the input width,
	the narrow MUXs lead to a smaller IM block.
With the smaller area, the table also shows lower power consumption.
The delay increased slightly, but the difference is negligible.
The seventh row shows the estimated Cambricon-X when the simplification
	proposed in Subsection \ref{subsec:acc} is applied.
The values were obtained by comparing the third and fifth rows.
Due to the area reduction of NBin, NBout, and IM,
	the total area can be reduced by 40\%.

The results of various configurations are presented at the last two rows
	for comparison.
With $N_{par}$=128, the areas of activation memory components are
	increased due to wide memory configuration.
The area of the IM block is not increased because
	the same $N_{group}$ leads to the same MUX input width.
The area of the IM block is affected by $N_{group}$ and $N_{mul}$.
The tables shows that $N_{mul}$ affects the area of SB
	though the capacity of SB is not changed.
This is also due to the memory configuration.

\section{Conclusions}
In this paper, an accelerator-aware pruning scheme was proposed for CNNs.
In the pruning process, the proposed scheme
	considers the accelerator parameters:
		the width of the internal activation buffer
		and the number of multipliers.
After pruning, each weight-fetching group
	has a fixed number of non-zero weights left.
Networks pruned by the proposed scheme can be efficiently run
	on the target accelerator.
Furthermore, the proposed pruning scheme can be used
	to reduce the complexity of the accelerators, too.
Even with the accelerator constraint, it was shown that
	the proposed scheme can reach pruning ratios close
	to those of existing unstructured pruning schemes.
In this paper, the pruning scheme was discussed
	in relation to representative sparse accelerator architectures,
	but the scheme can be used for any sparse architectures.

\ifCLASSOPTIONcaptionsoff
  \newpage
\fi

% trigger a \newpage just before the given reference
% number - used to balance the columns on the last page
% adjust value as needed - may need to be readjusted if
% the document is modified later
%\IEEEtriggeratref{8}
% The "triggered" command can be changed if desired:
%\IEEEtriggercmd{\enlargethispage{-5in}}

% references section

% can use a bibliography generated by BibTeX as a .bbl file
% BibTeX documentation can be easily obtained at:
% http://mirror.ctan.org/biblio/bibtex/contrib/doc/
% The IEEEtran BibTeX style support page is at:
% http://www.michaelshell.org/tex/ieeetran/bibtex/
%\bibliographystyle{IEEEtran}
% argument is your BibTeX string definitions and bibliography database(s)
%\bibliography{IEEEabrv,../bib/paper}
%
% <OR> manually copy in the resultant .bbl file
% set second argument of \begin to the number of references
% (used to reserve space for the reference number labels box)
\bibliographystyle{../../Templates/IEEE/IEEEtranBST_WIN_and_MAC_Bibliography/IEEEtran}
\bibliography{../../../references/Machine_Learning/cite_compressed}

%\begin{thebibliography}{1}
%
%\bibitem{IEEEhowto:kopka}
%H.~Kopka and P.~W. Daly, \emph{A Guide to \LaTeX}, 3rd~ed.\hskip 1em plus
%  0.5em minus 0.4em\relax Harlow, England: Addison-Wesley, 1999.
%
%\end{thebibliography}

% biography section
% 
% If you have an EPS/PDF photo (graphicx package needed) extra braces are
% needed around the contents of the optional argument to biography to prevent
% the LaTeX parser from getting confused when it sees the complicated
% \includegraphics command within an optional argument. (You could create
% your own custom macro containing the \includegraphics command to make things
% simpler here.)
%\begin{IEEEbiography}[{\includegraphics[width=1in,height=1.25in,clip,keepaspectratio]{mshell}}]{Michael Shell}
% or if you just want to reserve a space for a photo:

%\begin{IEEEbiography}{Michael Shell}
%Biography text here.
%\end{IEEEbiography}

% if you will not have a photo at all:
%\begin{IEEEbiographynophoto}{John Doe}
%Biography text here.
%\end{IEEEbiographynophoto}

% insert where needed to balance the two columns on the last page with
% biographies
%\newpage

%\begin{IEEEbiographynophoto}{Jane Doe}
%Biography text here.
%\end{IEEEbiographynophoto}

% You can push biographies down or up by placing
% a \vfill before or after them. The appropriate
% use of \vfill depends on what kind of text is
% on the last page and whether or not the columns
% are being equalized.

%\vfill

% Can be used to pull up biographies so that the bottom of the last one
% is flush with the other column.
%\enlargethispage{-5in}

% that's all folks
\end{document}